\documentclass{article}
\usepackage[utf8]{inputenc}
\usepackage{main}
\usepackage{microtype}
\usepackage{graphicx}
\usepackage{subfig}
\usepackage{times}
\usepackage{latexsym}
\usepackage{amsmath}
\usepackage{float}
\usepackage{footnote}
\usepackage{enumitem}
\usepackage{bm}
\usepackage{arydshln}
\usepackage{booktabs}
\usepackage{multicol}
\usepackage{multirow}
\usepackage{color}
\usepackage{xcolor}     
\usepackage{colortbl}
\usepackage{bbding}
\usepackage{makecell}
\usepackage{mathtools}
\usepackage{imakeidx}
\usepackage{longtable}
\usepackage{wrapfig}
\makeindex
\usepackage{arydshln}
\usepackage{lipsum}
\usepackage{natbib}
\usepackage[toc]{multitoc}
\usepackage[edges]{forest}
\usepackage[normalem]{ulem}
\definecolor{mydarkblue}{rgb}{0,0.08,0.45}
\usepackage[colorlinks=true,linkcolor=mydarkblue,citecolor=mydarkblue,filecolor=mydarkblue,urlcolor=mydarkblue]{hyperref}
\usepackage{caption}
\usepackage{CJKutf8}
\usepackage{awesomebox} % for infobox
\usepackage{bbding}
\usepackage[most]{tcolorbox}

\usepackage{graphicx}
\usepackage{array}
\usepackage{geometry}

\usepackage{amsmath}
\usepackage{amsfonts}
\usepackage{amssymb}

\usepackage[tikz]{bclogo}
\usepackage[framemethod=tikz]{mdframed}
\definecolor{bgblue}{RGB}{245,243,253}
\definecolor{ttblue}{RGB}{91,194,224}
\definecolor{mygray1}{gray}{.95}
\definecolor{mygray2}{gray}{.9}

\mdfdefinestyle{mystyle}{%
  rightline=true,
  innerleftmargin=10,
  innerrightmargin=10,
  outerlinewidth=3pt,
  topline=false,
  rightline=true,
  bottomline=false,
  skipabove=\topsep,
  skipbelow=\topsep
}

\newtcolorbox{myboxi}[1][]{
  breakable,
  title=#1,
  colback=red!5,
  colbacktitle=red!5,
  coltitle=black,
  fonttitle=\bfseries,
  bottomrule=0pt,
  toprule=0pt,
  leftrule=2pt,
  rightrule=2pt,
  titlerule=0pt,
  arc=0pt,
  outer arc=0pt,
  colframe=red,
}

\newtcolorbox{myboxnote}[1][]{
  breakable,
  title=#1,
  colback=orange!0,
  colbacktitle=orange!0,
  coltitle=black,
  fonttitle=\bfseries,
  bottomrule=0pt,
  toprule=0pt,
  leftrule=2pt,
  rightrule=2pt,
  titlerule=0pt,
  arc=0pt,
  outer arc=0pt,
  colframe=orange,
}

\newtcolorbox{myboxii}[1][]{
  breakable,
  freelance,
  title=#1,
  colback=white,
  colbacktitle=white,
  coltitle=black,
  fonttitle=\bfseries,
  bottomrule=0pt,
  boxrule=0pt,
  colframe=white,
  overlay unbroken and first={
  \draw[red!75!black,line width=3pt]
    ([xshift=5pt]frame.north west) -- 
    (frame.north west) -- 
    (frame.south west);
  \draw[red!75!black,line width=3pt]
    ([xshift=-5pt]frame.north east) -- 
    (frame.north east) -- 
    (frame.south east);
  },
  overlay unbroken app={
  \draw[red!75!black,line width=3pt,line cap=rect]
    (frame.south west) -- 
    ([xshift=5pt]frame.south west);
  \draw[red!75!black,line width=3pt,line cap=rect]
    (frame.south east) -- 
    ([xshift=-5pt]frame.south east);
  },
  overlay middle and last={
  \draw[red!75!black,line width=3pt]
    (frame.north west) -- 
    (frame.south west);
  \draw[red!75!black,line width=3pt]
    (frame.north east) -- 
    (frame.south east);
  },
  overlay last app={
  \draw[red!75!black,line width=3pt,line cap=rect]
    (frame.south west) --
    ([xshift=5pt]frame.south west);
  \draw[red!75!black,line width=3pt,line cap=rect]
    (frame.south east) --
    ([xshift=-5pt]frame.south east);
  },
}

\usepackage{fancyhdr} % to change header and footers
\usepackage{blindtext} % to quickly get a full document

\DeclareCaptionFont{black}{\color{black}}

\definecolor{myblue}{rgb}{0.9, 0.1, 0.94}
\definecolor{mygreen}{rgb}{0.64, 0.56, 0.88}
\definecolor{myyellow}{rgb}{0.68, 0.6, 0.1}
\definecolor{fancygreen}{rgb}{0.33, 0.68, 0.20}
\definecolor{salmon}{rgb}{0.94, 0.52, 0.49}
\definecolor{tablegreen}{rgb}{0.82, 0.94, 0.75}
\definecolor{tableblue}{rgb}{0.81, 0.90, 0.94}
\definecolor{tablered}{rgb}{0.97, 0.85, 0.85}
\definecolor{tableorange}{rgb}{0.96, 0.85, 0.81}

\newenvironment{itemize*}%
 {\leftmargini=10pt\begin{itemize}%
  \setlength{\itemsep}{0pt}%
  \setlength{\parskip}{0pt}%
  }%
 {\end{itemize}}
\newenvironment{enumerate*}%
 {\begin{enumerate}%
  \setlength{\itemsep}{0pt}%
  \setlength{\parskip}{0pt}}%
 {\end{enumerate}}

\usepackage{xcolor}
\usepackage{listings}

\newcommand\JSONnumbervaluestyle{\color{blue}}
\newcommand\JSONstringvaluestyle{\color{red}}

\newif\ifcolonfoundonthisline

\makeatletter

\lstdefinestyle{json}
{
  showstringspaces    = false,
  keywords            = {false,true},
  alsoletter          = 0123456789.,
  morestring          = [s]{"}{"},
  stringstyle         = \ifcolonfoundonthisline\JSONstringvaluestyle\fi,
  MoreSelectCharTable =%
    \lst@DefSaveDef{`:}\colon@json{\processColon@json},
  basicstyle          = \ttfamily,
  keywordstyle        = \ttfamily\bfseries,
}

\newcommand\processColon@json{%
  \colon@json%
  \ifnum\lst@mode=\lst@Pmode%
    \global\colonfoundonthislinetrue%
  \fi
}

\lst@AddToHook{Output}{%
  \ifcolonfoundonthisline%
    \ifnum\lst@mode=\lst@Pmode%
      \def\lst@thestyle{\JSONnumbervaluestyle}%
    \fi
  \fi
  \lsthk@DetectKeywords% 
}

\lst@AddToHook{EOL}%
  {\global\colonfoundonthislinefalse}

\makeatother

\usepackage{etoolbox}
\usepackage{natbib}
\usepackage{url}
\newcounter{bibcount}
\makeatletter
\patchcmd{\@lbibitem}{\item[}{\item[\hfil\stepcounter{bibcount}{[\thebibcount]}}{}{}
\renewcommand\NAT@bibsetup%
  [1]{\setlength{\leftmargin}{\bibhang}\setlength{\itemindent}{-\parindent}%
      \setlength{\itemsep}{\bibsep}\setlength{\parsep}{\z@}}
\makeatother

\author{%
  Zhen Huang$^2$, Zengzhi Wang$^{1, 2}$, Shijie Xia$^{1, 2}$, Pengfei Liu$^{1, 2}$
    \\
$^1$Shanghai Jiao Tong University, $^2$Generative AI Research Lab (GAIR)
\\
\texttt{gair.olympicarena@gmail.com}
}

\begin{document}

\title{OlympicArena Finals: Claude 3.5 Sonnet vs. GPT-4o} 
\title{OlympicArena Medal Ranks: \\GPT-4o vs. Claude 3.5 Sonnet vs. Gemini}
\title{
\raisebox{-0.2\height}{\includegraphics[width=0.08\textwidth]{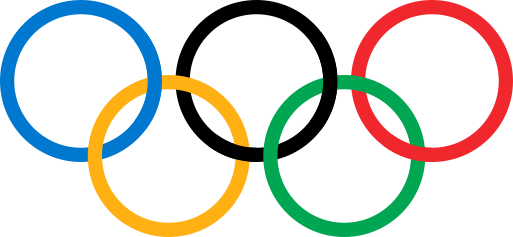}}
OlympicArena Medal Ranks: Who Is the Most Intelligent AI So Far?
}

\maketitle
\thispagestyle{fancy}
\fancyhead{}
\lhead{\includegraphics[height=0.67cm]{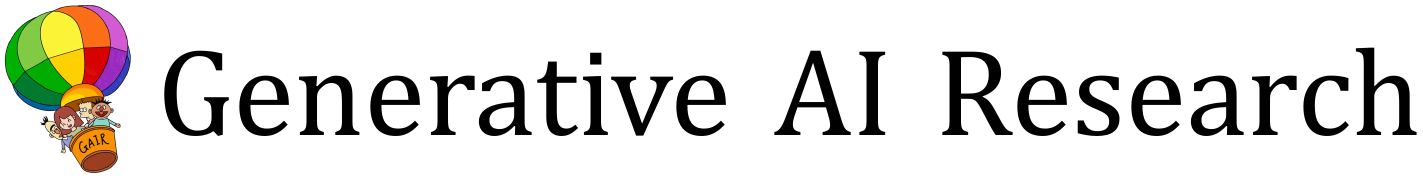}}
\renewcommand{\headrulewidth}{0pt}
\setlength{\headsep}{0mm}

\begin{abstract}

In this report, we pose the following question: \emph{\textbf{Who is the most intelligent AI model to date, as measured by the OlympicArena}}~\citep{huang2024olympicarena} (an  Olympic-level, multi-discipline, multi-modal benchmark for superintelligent AI)? We specifically focus on the most recently released models: ``\textbf{Claude-3.5-Sonnet}~\citep{anthropic_claude_2024},'' ``\textbf{Gemini-1.5-Pro}~\citep{reid2024gemini},'' and ``\textbf{GPT-4o}.'' 
For the first time, we propose using an \emph{Olympic medal Table} approach to rank AI models based on their comprehensive performance across various disciplines.
% Recently, \citet{huang2024olympicarena} has introduced ``\textit{OlympicArena},'' a benchmark designed to evaluate multi-discipline cognitive reasoning in superintelligent AI systems. With the emergence of the highly advanced model, \textbf{Claude-3.5-Sonnet}~\citep{anthropic_claude_2024} and \textbf{Gemini-1.5-Pro}~\citep{reid2024gemini}, we are keen to test their capabilities in the OlympicArena to ascertain their true potential. Additionally, this evaluation will determine if the latest models can surpass the current capabilities of GPT-4o (as well as GPT-4V). We conduct a thorough, apple-to-apple comparison to ensure a detailed assessment. 
Empirical results reveal:
\begin{itemize}
    \item Claude-3.5-Sonnet shows highly competitive overall performance over GPT-4o, even surpassing GPT-4o on a few subjects (i.e., \texttt{Physics}, \texttt{Chemistry} and \texttt{Biology}). 
    \item Gemini-1.5-Pro and GPT-4V are ranked consecutively just behind GPT-4o and Claude-3.5-Sonnet, but with a clear performance gap between them. 
    \item The performance of  AI models from the open-source community significantly lags behind these proprietary models. 
    \item The performance of these models on this benchmark has been less than satisfactory, indicating that we still have a long way to go before achieving superintelligence.
\end{itemize}

We remain committed to continuously tracking and evaluating the performance of the latest powerful models on this benchmark (available at \url{https://github.com/GAIR-NLP/OlympicArena}).

% \zzwang{
% Empirical results reveal that Claude-3.5-Sonnet shows highly competitive overall performance over GPT-4o, even surpassing GPT-4o on a few subjects. Meanwhile, Gemini-1.5-Pro and GPT-4V are ranked consecutively just behind GPT-4o and Claude-3.5-Sonnet, but with a clear performance gap between them. Furthermore, the performance of  AI models from the open-source community significantly lags behind these proprietary models. In conclusion, the performance of these models on this benchmark has been less than satisfactory, indicating that we still have a long way to go before achieving superintelligence. We remain committed to continuously tracking and evaluating the performance of the latest powerful models on this benchmark.
% }

\end{abstract}

% \pfliu{We need an olympic-style leaderboard below.}

\vspace{0.9cm}

% Title section with medals and header
% \begin{center}
    
% \end{center}

% \begin{center}
%     \Huge \textbf{Leaderboard}
% \end{center}

% Table with medal counts
\begin{center}
\vspace{-0.79cm}

\includegraphics[width=0.550\textwidth]{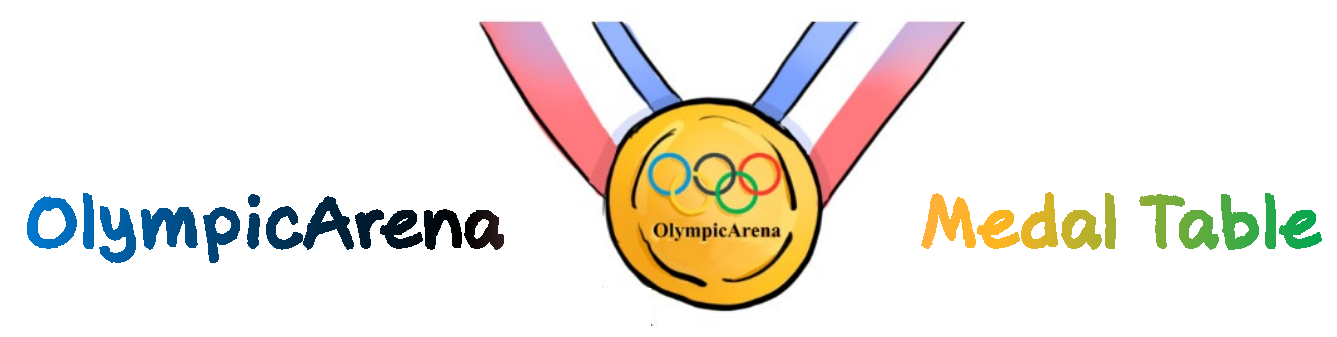}

\renewcommand{\arraystretch}{1.5}
\scalebox{0.75}{
\begin{tabular}{clccccc}
    \toprule
    \rowcolor{yellow!20}
    % \rowcolor[HTML]{EA6B6F}
    \raisebox{-0.2\height}{\includegraphics[width=0.05\textwidth]{fig/Olympic_rings_without_rims.svg.png}}
    
     \textbf{Ranking} & \textbf{Models} &
         
     \raisebox{-0.2\height}{\includegraphics[width=0.03\textwidth]{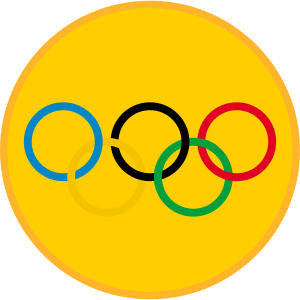}}
     \textbf{Gold} & \raisebox{-0.2\height}{\includegraphics[width=0.03\textwidth]{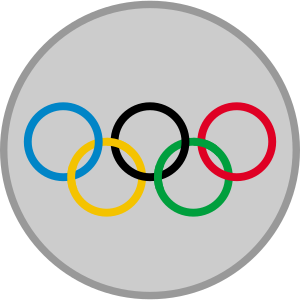}} \textbf{Sliver} & \raisebox{-0.2\height}{\includegraphics[width=0.03\textwidth]{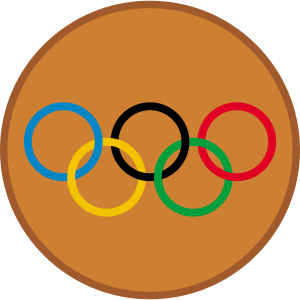}} \textbf{Bronze} & \textbf{Total} &\textbf{Overall Scores}  \\
    \midrule
    \rowcolor{mygray1} 1 & GPT-4o & \textbf{4} & \textbf{3} & 0 & \textbf{7} & 40.47 \\
    2 & Claude-3.5-Sonnet & \textbf{3} & \textbf{3} & 0 & \textbf{6} & 39.24 \\
    
    \rowcolor{mygray1} 4 & GPT-4V & 0 & \textbf{1} & \textbf{1} & \textbf{2} & 33.17 \\
    3 & Gemini-1.5-Pro & 0 & 0 & \textbf{6} & \textbf{6} & 35.09 \\

    \rowcolor{mygray1} 5 & Claude-3-Sonnet & 0 & 0 & 0 & 0 & 25.53 \\
    6 & Qwen1.5-32B-Chat & 0  & 0 & 0 & 0 & 24.36 \\
    \rowcolor{mygray1} 7 & Qwen-VL-Max & 0  & 0 & 0 & 0 & 21.41 \\
    8 & Gemini-Pro-Vision & 0  & 0 & 0 & 0 & 21.02 \\
    \rowcolor{mygray1} 9 & LLaVA-NeXT-34B & 0  & 0 & 0 & 0 & 18.16 \\
    10 & Yi-34B-Chat& 0  & 0 & 0 & 0 & 18.01 \\
    \rowcolor{mygray1} 11 & InternVL-Chat-V1.5 & 0  & 0 & 0 & 0 & 17.39 \\
    12 & InternLM2-Chat-20B & 0 & 0 & 0 & 0 & 17.33 \\
    \rowcolor{mygray1} 13 & Yi-VL-34B & 0 & 0 & 0 & 0 & 15.07 \\
    14 & Qwen-VL-Chat & 0 & 0 & 0 & 0 & 7.34 \\
    \rowcolor{mygray1} 15 & Qwen-7B-Chat & 0 & 0 & 0 & 0 & 4.34 \\
    \bottomrule
\end{tabular}
}
\captionof{table}{The medal table of various models across disciplines (June 23, 2024). Note that we rank AI models by Gold medals first, then by Silver medals, then by Bronze medals, and finally by Overall score if tied.}
\label{medal_table}
\end{center}

\newpage

\pagestyle{fancy}
\lhead{\rightmark}
\renewcommand{\headrulewidth}{0.7pt}
\setlength{\headsep}{5mm}

% \tableofcontents

\clearpage

\section{OlympicArena}

Recently, \citet{huang2024olympicarena} introduce the \textit{OlympicArena}, including 11,163 bilingual problems
across both text-only and interleaved text-image modalities and spanning seven common subjects and 62 international
Olympic competitions, rigorously examined for data leakage. As one of the most comprehensive and challenging benchmarks available, OlympicArena aims to push the boundaries of AI capabilities in cognitive reasoning, requiring models to demonstrate proficiency across diverse and complex problem sets. In this report, we present several key contributions:
\begin{enumerate}
    \item \textbf{Comparison of the Latest Models}: We analyze and compare two newly released advanced models from the past month—Claude-3.5-Sonnet, Gemini-1.5-Pro, against OpenAI's GPT series. This comparison provides valuable insights into the performance of these cutting-edge models.
    \item \textbf{Invention of the OlympicArena Medal Table}: We have created a novel ranking mechanism, the OlympicArena Medal Table, which offers a clear and competitive framework for comparing different models. 
    \item \textbf{Detailed Fine-Grained Analysis}: We conduct a fine-grained analysis of the latest and most powerful models, enhancing the OlympicArena benchmark by providing a deeper understanding of model capabilities and limitations.
    
\end{enumerate}

% \textbf{(1) Comparison of the Latest Models:} We analyze and compare two newly released advanced models from the past month—Claude-3.5-Sonnet, Gemini-1.5-Pro, against OpenAI's GPT series. This comparison provides valuable insights into the performance of these cutting-edge models.

% \textbf{(2) Invention of the OlympicArena Medal Table:} We have created a novel ranking mechanism, the OlympicArena Medal Table, which offers a clear and competitive framework for comparing different models. 

% \textbf{(3) Detailed Fine-Grained Analysis:} We conduct a fine-grained analysis of the latest and most powerful models, enhancing the OlympicArena benchmark by providing a deeper understanding of model capabilities and limitations.

\subsection{Setup}

In this report, we use the test split of the \textit{OlympicArena} dataset. The answers for this split are not publicly available, which helps to prevent data leakage and reflects the true performance of the model. Additionally, the \textit{OlympicArena-test} split does not include data that requires model-based evaluation; all evaluations can be performed using rule-based matching. This report also tests both Large Multimodal Models (LMMs) and Large Language Models (LLMs). For the LLM test, we use the text-only setting, providing no any image-related information to the model during input. We do not use image captions as textual representations of images because, in research with \textit{OlympicArena}, it turns out that image captions are not always effective. The role of image captions may require further exploration.
%\zzwang{The role of image captions may require further exploration.}
Therefore, to maintain the original structure of the problems, we directly use text-only inputs. We utilize the zero-shot CoT prompt consistent with the \textit{OlympicArena} paper.

\subsection{Competitors}

We evaluate a range of both open-source and proprietary LMMs and LLMs. For LMMs, we select GPT-4o~\citep{gpt4o}, GPT-4V~\citep{GPT4Vision}, Claude-3-Sonnet~\citep{claude}, Gemini Pro Vision~\citep{geminiPro}, Qwen-VL-Max~\citep{qwen-vl}. We also evaluate several open-source models, including LLaVA-NeXT-34B~\citep{llava-next}, InternVL-Chat-V1.5~\citep{chen2023internvl}, Yi-VL-34B~\citep{yi}, and Qwen-VL-Chat~\citep{qwen-vl}. For LLMs, we select open-source models like Qwen-7B-Chat, Qwen1.5-32B-Chat~\citep{qwen}, Yi-34B-Chat~\citep{yi}, and InternLM2-Chat-20B~\citep{internlm2}. 

Moreover, \textbf{we particularly include the newly released Claude-3.5-Sonnet~\citep{anthropic_claude_2024} as well as Gemini-1.5-Pro~\citep{geminiteam2024gemini}, and compare them with the powerful GPT-4o and GPT-4V.} This comparison allows us to evaluate the latest advancements in model capabilities and performance.

\subsection{Evaluation Methods}

\paragraph{Metrics:}

Since all problems can be evaluated using rule-based matching, we use accuracy for non-programming tasks and unbiased pass@k for programming tasks, defined as follows:
\begin{equation}
\label{eq:pass-at-k}
\operatorname{pass} @ k := \underset{\text {Problems}}{\mathbb{E}}\left[1-\frac{\binom{n-c}{k}}{\binom{n}{k}}\right]
\end{equation}
where we set \( k=1 \) and \( n=5 \), and \( c \) indicates the number of correct samples that pass all test cases.

\paragraph{OlympicArena Medal Table:}

The \emph{OlympicArena Medal Table}, similar to the medal system used in the Olympic Games,~\footnote{\url{https://olympics.com/}}
% \zzwang{, similar to the medal system used in the Olympic Games,}
is a pioneering ranking mechanism specifically designed to evaluate the performance of AI models across various academic disciplines. This table awards medals to models that achieve the top three scores in any given discipline, thereby providing a clear and competitive framework for comparing different models. 
%\zzwang{Specifically, we rank AI models by Gold medals first, then by overall scores if tied.}
Specifically, we rank AI models by Gold medals first, then by Silver medals, then by Bronze medals, and finally by Overall score if tied. It offers a straightforward and intuitive way to identify leading models in distinct academic fields, making it easier for researchers and developers to understand the strengths and weaknesses of different models.

\paragraph{Fine-grained Evaluation:}

We conduct a fine-grained evaluation based on accuracy across different disciplines, modalities, languages, as well as different types of logical and visual reasoning abilities.

\section{Results and Analysis}

It should be noted that this report primarily focuses on the analysis of experimental results comparing top-performing Claude-3.5-Sonnet and GPT-4o, while also providing some commentary on the performance of Gemini-1.5-Pro.

\subsection{Overall}

From the overall results in Table~\ref{results_subject}, it can be observed that:
\begin{itemize*}
    \item the newly released Claude-3.5-Sonnet is very powerful, reaching a level \textbf{almost on par with GPT-4o}. The difference in overall accuracy between the two is only about 1\%. 
    \item At the same time, the newly released Gemini-1.5-Pro also demonstrates considerable strength, surpassing GPT-4V (OpenAI's current second most powerful model) in most disciplines. 
    \item Additionally, according to the OlympicArena Medal Table (see Table~\ref{medal_table}), where a model earns a medal if it achieves one of the top three scores in any discipline, we can observe that \textbf{GPT-4o, Claude-3.5-Sonnet, and Gemini-1.5-Pro are the top three models in the ranking}.
    \item It is worth noting that at the time of writing this report, \textbf{the oldest of these three models has been released for only about a month}, reflecting the rapid development of this field.
    \item Based on the medal table, we have also discovered a \textbf{significant gap between open-source and proprietary models} as open-source models have not managed to secure a medal in any discipline.
\end{itemize*}

% (1) the newly released Claude-3.5-Sonnet is very powerful, reaching a level \textbf{almost on par with GPT-4o}. The difference in overall accuracy between the two is only about 1\%. 

% (2) At the same time, the newly released Gemini-1.5-Pro also demonstrates considerable strength, surpassing GPT-4V (OpenAI's current second most powerful model) in most disciplines. 

% (3) Additionally, according to the OlympicArena Medal Table (see Table~\ref{medal_table}), where a model earns a medal if it achieves one of the top three scores in any discipline, we can observe that \textbf{GPT-4o, Claude-3.5-Sonnet, and Gemini-1.5-Pro are the top three models in the ranking}.

% (4) It is worth noting that at the time of writing this paper, \textbf{the oldest of these three models has been released for only about a month}, reflecting the rapid development of this field.

% (5) Based on the medal table, we have also discovered a \textbf{significant gap between open-source and proprietary models} as open-source models have not managed to secure a medal in any discipline.

% \zzwang{use a short sentence to describe the performance of open-source models.} \zzwang{describe the medal table in the first page.}

% \subsection{GPT-4o Vs Claude-3.5-Sonnet across Various Disciplines}

% Table generated by Excel2LaTeX from sheet 'Sheet1'
\begin{table}[htbp]
  \centering
    \caption{Experimental results across different subjects on OlympicArena benchmark, expressed as percentages, with the highest score bolded. We use the pass@k metric (Equation~\ref{eq:pass-at-k}) for CS problems. When calculating the overall accuracy, for code generation problems, if any generated code for a problem passes all test cases, the problem is considered correct.}
  \scalebox{0.8}{
    \begin{tabular}{l|cccccccc}
    \toprule
    \multicolumn{1}{c|}{} &  \textbf{Math} & \textbf{Physics} & \textbf{Chemistry} & \textbf{Biology} & \textbf{Geography} & \textbf{Astronomy} & \textbf{CS} & \textbf{Overall} \\ \cmidrule(lr){2-9}
\multicolumn{1}{l|}{\multirow{-2}{*}{Model}}  & Accuracy & Accuracy & Accuracy & Accuracy & Accuracy & Accuracy & Pass@1 & Accuracy \\ 
    % Model & Setting & \multicolumn{1}{l}{Math} & \multicolumn{1}{l}{Physics} & \multicolumn{1}{l}{Chemistry} & \multicolumn{1}{l}{Biology} & \multicolumn{1}{l}{Geography} & \multicolumn{1}{l}{Astronomy} & \multicolumn{1}{l}{CS} & \multicolumn{1}{l}{Overall} \\
    \midrule
    GPT-4o & \textbf{28.32}  & 30.01  & 46.68  & 53.11  & \textbf{56.77}  & \textbf{44.50}  & \textbf{8.43}  & \textbf{40.47}  \\
    Claude-3.5-Sonnet & 23.18  & \textbf{31.16}  & \textbf{47.27}  & \textbf{56.05}  & 55.19  & 43.51  & 5.19  & 39.24  \\
    Gemini-1.5-Pro & 19.99 & 28.93 & 43.80 & 49.16 & 49.67 & 38.29 & 5.37 & 35.09 \\
    \midrule
    GPT-4V & 18.98  & 24.94  & 41.06  & 47.69  & 50.33  & 32.07  & 6.94  & 33.17  \\
    Claude3 Sonnet & 7.12  & 18.42  & 30.06  & 39.40  & 40.80  & 24.50  & 1.02  & 25.53  \\
    Qwen1.5-32B-Chat & 9.41  & 15.81  & 32.13  & 39.00  & 40.41  & 27.84  & 0.28  & 24.36  \\
    Qwen-VL-Max & 6.68  & 14.12  & 24.52  & 36.32  & 40.41  & 23.42  & 0.83  & 21.41  \\
    Gemini Pro Vision & 5.91  & 12.97  & 28.36  & 37.66  & 37.71  & 20.54  & 1.39  & 21.02  \\
    LLaVA-NeXT-34B & 2.99  & 11.59  & 22.38  & 33.44  & 36.99  & 18.47  & 0.19  & 18.16  \\
    Yi-34B-Chat & 3.06  & 11.13  & 24.52  & 33.18  & 34.69  & 18.29  & 0.19  & 18.01  \\
    InternVL-Chat-V1.5 & 6.08  & 10.82  & 20.09  & 30.57  & 33.18  & 15.95  & 0.19  & 17.39  \\
    Internlm2-Chat-20B & 5.78  & 11.13  & 19.13  & 31.91  & 32.13  & 16.49  & 0.46  & 17.33  \\
    Yi-VL-34B & 2.92  & 10.82  & 21.05  & 28.09  & 25.16  & 16.94  & 0.00  & 15.07  \\
    Qwen-VL-Chat & 1.71  & 5.22  & 9.08  & 12.44  & 14.06  & 8.11  & 0.00  & 7.34  \\
    Qwen-7B-Chat & 1.55  & 4.45  & 6.50  & 7.29  & 4.60  & 5.59  & 0.00  & 4.34  \\

    \bottomrule
    \end{tabular}}
  \label{results_subject}%
\end{table}%

\subsection{Fine-grained Analysis w.r.t Subject}

\paragraph{GPT-4o vs. Claude-3.5-Sonnet}

As seen in Table~\ref{results_subject}, although GPT-4o and Claude-3.5-Sonnet have similar overall performance across disciplines, each model exhibits specific strengths. GPT-4o demonstrates superior capabilities in traditional deductive and inductive reasoning tasks, particularly in mathematics and computer science, outperforming Claude-3.5-Sonnet by over 5\% in mathematics and 3\% in computer science. On the other hand, Claude-3.5-Sonnet excels in subjects such as physics, chemistry, and biology, especially in biology where it surpasses GPT-4o by 3\%. 

\paragraph{GPT-4V vs. Gemini-1.5-Pro} A similar pattern can be observed in the comparison between Gemini-1.5-Pro and GPT-4V. Gemini-1.5-Pro performs significantly better than GPT-4V in physics, chemistry, and biology. However, in mathematics and computer science, Gemini-1.5-Pro shows only a slight advantage or even a disadvantage.

\paragraph{Insights}
From these two sets of comparisons, we can infer that:
\begin{enumerate}
    \item OpenAI's GPT series remains exceptionally well-tuned for traditional mathematical reasoning and coding abilities. This superior performance in these two subjects suggests that GPT models have been rigorously trained to handle tasks that require strong deductive reasoning and algorithmic thinking.
    \item Conversely, when it comes to disciplines that require the integration of knowledge with reasoning, such as physics, chemistry, and biology, other models like Claude-3.5-Sonnet and Gemini-1.5-Pro demonstrate competitive or superior performance. This observation highlights the distinct areas of expertise and potential training focuses for different model series, indicating a possible trade-off between specialization in reasoning-intensive tasks and broader knowledge integration.
\end{enumerate}

% OpenAI's GPT series remains exceptionally well-tuned for traditional mathematical reasoning and coding abilities. This superior performance in these two subjects suggests that GPT models have been rigorously trained to handle tasks that require strong deductive reasoning and algorithmic thinking. Conversely, when it comes to disciplines that require the integration of knowledge with reasoning, such as physics, chemistry, and biology, other models like Claude-3.5-Sonnet and Gemini-1.5-Pro demonstrate competitive or superior performance. This observation highlights the distinct areas of expertise and potential training focuses for different model series, indicating a possible trade-off between specialization in reasoning-intensive tasks and broader knowledge integration.

% Table generated by Excel2LaTeX from sheet 'Logical'
\begin{table}[htbp]
  \centering
  \caption{Performance of various models on logical easoning abilities. Logical reasoning  abilities: Deductive Reasoning (DED), Inductive Reasoning (IND), Abductive Reasoning (ABD),  Analogical Reasoning (ANA), Cause-and-Effect Reasoning (CAE), Critical Thinking (CT), Decompositional Reasoning (DEC), and Quantitative Reasoning (QUA).}
  \label{results_logic}%
  \scalebox{0.9}{
    \begin{tabular}{lcccccccc}
    \toprule
    \textbf{Model} & \textbf{DED} & \textbf{IND} & \textbf{ABD} & \textbf{ANA} & \textbf{CAE} & \textbf{CT} & \textbf{DEC} & \textbf{QUA} \\
    \midrule

    GPT-4o & \textbf{42.26}  & \textbf{32.26}  & \textbf{51.27}  & \textbf{42.46}  & 46.74  & \textbf{47.97}  & 33.78  & 38.27  \\
    Claude-3.5-Sonnet & 41.80  & 31.85  & 50.91  & 41.05  & \textbf{47.01}  & 47.61  &\textbf{33.95}  & \textbf{38.38}  \\
    Gemini-1.5-Pro & 37.83 & 28.21  & 47.64 & 35.79 & 42.26 & 43.09 & 30.62 & 34.60 \\
    GPT-4V & 35.40  & 25.14  & 47.82  & 34.04  & 40.55  & 42.03  & 26.75  & 31.10  \\
    Qwen1.5-32B-chat & 27.23  & 21.75  & 34.00  & 25.03  & 34.10  & 33.09  & 21.37  & 23.38  \\
    Claude3-Sonnet & 26.56  & 19.48  & 37.45  & 24.68  & 31.00  & 32.88  & 19.23  & 23.32  \\
    Qwen-VL-Max & 24.47  & 17.38  & 35.45  & 21.87  & 31.22  & 29.94  & 18.48  & 19.18  \\
    Gemini Pro Vision & 23.41  & 17.38  & 36.00  & 21.99  & 26.95  & 29.20  & 17.64  & 19.33  \\
    LLaVA-NeXT-34B & 21.26  & 14.96  & 31.64  & 20.70  & 29.19  & 26.91  & 14.63  & 15.89  \\
    Yi-34B-Chat & 20.99  & 14.47  & 26.00  & 19.53  & 27.91  & 26.97  & 16.11  & 16.63  \\
    InternVL-Chat-V1.5 & 19.00  & 13.34  & 31.45  & 18.48  & 23.48  & 24.18  & 13.48  & 15.45  \\
    Internlm2-Chat-20B & 18.81  & 13.66  & 25.64  & 17.54  & 24.17  & 23.91  & 14.05  & 14.44  \\
    Yi-VL-34B & 17.33  & 10.91  & 20.91  & 16.84  & 22.79  & 21.80  & 12.52  & 14.21  \\
    Qwen-VL-Chat & 8.54  & 6.71  & 13.27  & 9.24  & 10.51  & 10.81  & 5.78  & 6.66  \\
    Qwen-7B-Chat & 4.95  & 4.69  & 5.64  & 5.61  & 5.39  & 5.29  & 4.31  & 4.59  \\
    \bottomrule
    \end{tabular}%
    }
\end{table}%

% Table generated by Excel2LaTeX from sheet 'Visual'
% \begin{table}[htbp]
\begin{wraptable}{r}{8cm}
  \centering
  \caption{Performance of various models on visual reasoning abilities. Visual reasoning abilities: Pattern  Recognition (PR), Spatial Reasoning (SPA), Diagrammatic Reasoning (DIA), Symbol Interpretation  (SYB), and Comparative Visualization (COM). We bold the highest scores.}
  \label{results_visual}%
  
   \resizebox{0.5\textwidth}{!}{
    \begin{tabular}{lccccc}
    \toprule
    \textbf{Model} & \textbf{PR} & \textbf{SPA} & \textbf{DIA} & \textbf{SYB} & \textbf{COM} \\
    \midrule
    GPT-4o & 41.88  & \textbf{31.64}  & 38.01  & \textbf{34.64}  & \textbf{41.31}  \\
    Claude-3.5-Sonnet & \textbf{42.30}  & 29.99  & \textbf{38.21}  & \textbf{34.64}  & 40.97  \\
    Gemini-1.5-Pro & 37.20  & 26.95  & 32.95  & 29.88  & 36.02\\
    GPT-4V & 35.96  & 24.00  & 30.81  & 27.61  & 35.07  \\
    Qwen1.5-32B-chat & 29.38  & 18.46  & 25.61  & 23.17  & 28.29  \\
    Gemini Pro Vision & 29.29  & 15.42  & 22.31  & 20.09  & 26.64  \\
    Qwen-VL-Max & 28.18  & 17.22  & 23.32  & 20.40  & 26.72  \\
    Claude3-Sonnet & 27.35  & 18.01  & 22.83  & 21.02  & 25.82  \\
    LLaVA-NeXT-34B & 24.85  & 13.77  & 20.35  & 16.95  & 23.21  \\
    Yi-34B-Chat & 23.97  & 14.47  & 20.35  & 18.44  & 22.47  \\
    InternVL-Chat-V1.5 & 23.32  & 13.07  & 18.67  & 15.95  & 21.77  \\
    Internlm2-Chat-20B & 23.04  & 13.97  & 19.51  & 16.48  & 21.52  \\
    Yi-VL-34B & 16.89  & 9.63  & 13.93  & 12.41  & 15.24  \\
    Qwen-VL-Chat & 10.37  & 5.39  & 7.98  & 6.90  & 9.38  \\
    Qwen-7B-Chat & 4.72  & 2.69  & 4.42  & 4.10  & 4.75  \\
    \bottomrule
    \end{tabular}%
    }
% \end{table}%
\end{wraptable}

\subsection{Fine-grained Analysis w.r.t Reasoning Type}

\paragraph{GPT-4o vs. Claude-3.5-Sonnet on different logical reasoning abilities}

Table~\ref{results_logic} presents the performance of various models on different logical reasoning abilities. The comparison between GPT-4o and Claude-3.5-Sonnet reveals several insights. GPT-4o generally outperforms Claude-3.5-Sonnet in most logical reasoning abilities, notably Deductive Reasoning, Inductive Reasoning, Abductive Reasoning, Analogical Reasoning, and Critical Thinking. However, Claude-3.5-Sonnet surpasses GPT-4o in Cause-and-Effect Reasoning, Decompositional Reasoning, and Quantitative Reasoning. Despite these slight differences, both models exhibit relatively comparable performance across the board, indicating that while GPT-4o has a slight edge in most categories, Claude-3.5-Sonnet remains competitive, particularly in reasoning tasks that involve cause-and-effect analysis and decomposition.

\paragraph{GPT-4o vs. Claude-3.5-Sonnet on different visual reasoning abilities}

Table~\ref{results_visual} presents the performance of various models on different visual reasoning abilities. Comparing GPT-4o and Claude-3.5-Sonnet, we observe that Claude-3.5-Sonnet leads in Pattern Recognition and Diagrammatic Reasoning, indicating its strength in identifying patterns and interpreting diagrams. Both models perform equally well in Symbol Interpretation, suggesting a comparable capability in understanding and processing symbolic information. However, GPT-4o outperforms Claude-3.5-Sonnet in Spatial Reasoning and Comparative Visualization, demonstrating its superiority in tasks that require understanding spatial relationships and comparing visual data.

\paragraph{Insight: Relationship between disciplines and reasoning abilities}

\begin{enumerate}
    \item Analyzing the correlation between performance across subjects and reasoning abilities, we can infer that mathematics and computer programming, which emphasize general complex deductive reasoning skills and the derivation of universal conclusions based on rules, tend to \textbf{rely less on extensive pre-existing knowledge}. In contrast, subjects like chemistry and biology often \textbf{require a substantial knowledge base} to make inferences based on known information about causality and phenomena. This suggests that while mathematical and coding abilities remain effective measures of a model's reasoning capabilities, \textbf{other subjects better test a model's ability to assist with reasoning and problem analysis based on its internal knowledge}. 
    \item Moreover, this analysis highlights the \textbf{importance of tailored training datasets that encompass a broad spectrum of knowledge domains}. For instance, to improve performance in knowledge-intensive subjects like chemistry and biology, models need extensive exposure to domain-specific data during training. Conversely, for disciplines requiring robust logical and deductive reasoning, such as mathematics and computer science, \textbf{models benefit from training that focuses on pure logical reasoning and novel reasoning frameworks}.
    \item Additionally, \textbf{the distinction between reasoning abilities and knowledge application underscores the potential for interdisciplinary applications of these models}. For example, a model with strong deductive reasoning capabilities can assist in fields that require systematic problem-solving, such as scientific research and engineering. Meanwhile, a model with a rich knowledge base can be invaluable in disciplines that rely heavily on existing information, such as medicine and environmental science. Understanding these nuances not only aids in the development of more specialized and versatile models but also emphasizes the need for continuous evaluation and refinement of model architectures to better align with the diverse requirements of different academic and professional fields.
\end{enumerate}

\subsection{Fine-grained Analysis w.r.t Language Type}

% Table generated by Excel2LaTeX from sheet 'Sheet2'
% \begin{table}[htbp]
\begin{wraptable}{r}{6cm}
  \centering
  \caption{Experimental results across different languages (English and Chinese) on OlympicArena benchmark, expressed as percentages, with the highest score bolded.}
\scalebox{0.88}{
\begin{tabular}{l|cc}
    \toprule
        \multicolumn{1}{c|}{}  & \textbf{EN} & \textbf{ZH}\\ \cmidrule(lr){2-3}
\multicolumn{1}{l|}{\multirow{-2}{*}{Model}}  & Accuracy & Accuracy  \\ 
\midrule
    GPT-4o & \textbf{44.16}  & \textbf{34.59}  \\
    Claude-3.5-Sonnet & 43.09  & 32.83  \\
    Gemini-1.5-Pro & 38.58  & 29.27  \\
    GPT-4V & 37.17  & 26.49  \\
    Claude-3-Sonnet & 27.41  & 17.08  \\
    Qwen1.5-32B-Chat & 23.80  & 25.29  \\
    Qwen-VL-Max & 21.34  & 21.52  \\
    Gemini Pro Vision & 22.48  & 18.58  \\
    LLaVA-NeXT-34B & 19.00  & 16.76  \\
    Yi-34B-Chat & 17.43  & 18.98  \\
    InternVL-Chat-V1.5 & 18.36  & 15.77  \\
    Internlm2-Chat-20B & 17.82  & 16.52  \\
    Yi-VL-34B & 15.20  & 14.86  \\
    Qwen-VL-Chat & 8.32  & 5.69  \\
    Qwen-7B-Chat & 4.20  & 4.57  \\
    \bottomrule
    \end{tabular}%
}
  \label{results_language}%
% \end{table}%
\end{wraptable}

Table~\ref{results_language} presents a comparison of model performance across different languages. We find that most models have \textbf{higher accuracy in English compared to Chinese}, and this disparity is particularly significant among the top-ranked models. We speculate that there are several reasons for this: 
\begin{enumerate*}
    \item These models are still primarily trained on English data, despite including some Chinese data and having cross-linguistic generalization capabilities.
    \item The difficulty of Chinese problems is more challenging than that of English problems, especially in subjects like physics and chemistry where Chinese Olympiad problems are notoriously harder.
    \item The models still have poor support for recognizing characters in multimodal images, which is more severe in Chinese.
\end{enumerate*}
% (1) These models are still primarily trained on English data, despite including some Chinese data and having cross-linguistic generalization capabilities; (2) The difficulty of Chinese problems is more challenging than that of English problems, especially in subjects like physics and chemistry where Chinese Olympiad problems are notoriously harder; (3) The models still have poor support for recognizing characters in multimodal images, which is more severe in Chinese.

However, we also find that some models either developed in China or fine-tuned on base models supporting Chinese \textbf{perform better in Chinese scenarios than in English ones}. Examples include Qwen1.5-32B-Chat, Qwen-VL-Max, Yi-34B-Chat, and Qwen-7B-Chat. Other models such as InternLM2-Chat-20B and Yi-VL-34B, although still performing better in English, show much smaller accuracy differences between English and Chinese scenarios compared to the top-ranked closed-source models. This indicates that \textbf{optimizing models for Chinese data, and for more languages worldwide, still requires significant attention}.

% Table generated by Excel2LaTeX from sheet 'Sheet3'
\begin{table}[htbp]
% \begin{wraptable}{l}{6cm}
  \centering
  \caption{Experimental results across different modalities (text-only and multi-modal) on OlympicArena benchmark, expressed as percentages, with the highest score bolded.}
  \label{results_modality}%
    \scalebox{0.8}{
    \begin{tabular}{l|cc}
    \toprule
    \multicolumn{1}{c|}{}  & \textbf{Text-only} & \textbf{Multi-modal}\\ \cmidrule(lr){2-3}
\multicolumn{1}{l|}{\multirow{-2}{*}{Model}}  & Accuracy & Accuracy  \\ 
    \midrule
    GPT-4o & \textbf{41.79}  & \textbf{39.03}  \\
    Claude-3.5-Sonnet & 39.64  & 38.73  \\
    Gemini-1.5-Pro & 35.77  & 34.23  \\
    GPT-4V & 33.96  & 32.16  \\
    Claude-3-Sonnet & 22.80  & 24.47  \\
    Qwen1.5-32B-Chat & 23.57  & 25.35  \\
    Qwen-VL-Max & 19.86  & 23.37  \\
    Gemini Pro Vision & 19.38  & 23.10  \\
    LLaVA-NeXT-34B & 16.39  & 20.41  \\
    Yi-34B-Chat & 16.26  & 20.23  \\
    InternVL-Chat-V1.5 & 15.85  & 19.34  \\
    Internlm2-Chat-20B & 15.76  & 19.32  \\
    Yi-VL-34B & 16.48  & 13.29  \\
    Qwen-VL-Chat & 6.52  & 8.38  \\
    Qwen-7B-Chat & 4.62  & 3.98  \\
    \bottomrule
    \end{tabular}%
    }
\end{table}%
% \end{wraptable}

\subsection{Fine-grained Analysis w.r.t Modality}
% \paragraph{Text-only vs. Multimodal}

As shown in Table~\ref{results_modality}, GPT-4o outperforms Claude-3.5-Sonnet in both text-only and multi-modal tasks, particularly excelling in text-only problems. The gap between the two models is slightly larger in text-only tasks compared to multi-modal ones. On the other hand, Gemini-1.5-Pro performs better over GPT-4V in both types of problems. These observations indicate that even \textbf{the most powerful models currently available achieve higher accuracy in text-only tasks compared to multi-modal ones}. Although the difference is not substantial, it suggests that there is still \textbf{considerable room for improvement in models' ability to utilize multi-modal information} to tackle complex reasoning problems.

\section{Conclusion}

In this report, we primarily focus on the latest powerful models: Claude-3.5-Sonnet and Gemini-1.5-Pro, and compare them with OpenAI's GPT-4o and GPT-4V. We have also designed a novel ranking system for large models, the OlympicArena Medal Table, which provides a clear and competitive framework for comparing different models.
We find that GPT-4o, compared to its competitors, excels in subjects like mathematics and computer science, which correspond to complex deductive reasoning skills and the derivation of universal conclusions based on rules. On the other hand, Claude-3.5-Sonnet is better at making inferences based on known information about causality and phenomena. Additionally, we observe that these powerful models still perform better on English language problems and have significant room for improvement in their multi-modal capabilities.
Understanding these nuances aids in developing more specialized and versatile models and highlights the importance of continuous evaluation and refinement of model architectures to better meet the diverse requirements of different academic and professional fields.

% \zzwang{several summary sentences, then highlight the  takeaways, outlook, and limitations}

% \paragraph{Takeaways}

% \paragraph{Outlook}

% \paragraph{Limitations}

\clearpage
\bibliography{misc}

\begin{thebibliography}{14}
\expandafter\ifx\csname natexlab\endcsname\relax\def\natexlab#1{#1}\fi

\bibitem[{GPT(2023)}]{GPT4Vision}
 2023.
\newblock \href {https://api.semanticscholar.org/CorpusID:263218031} {Gpt-4v(ision) system card}.

\bibitem[{Anthropic(2024{\natexlab{a}})}]{anthropic_claude_2024}
Anthropic. 2024{\natexlab{a}}.
\newblock Claude 3.5 sonnet.
\newblock \url{https://www.anthropic.com/news/claude-3-5-sonnet}.

\bibitem[{Anthropic(2024{\natexlab{b}})}]{claude}
AI~Anthropic. 2024{\natexlab{b}}.
\newblock The claude 3 model family: Opus, sonnet, haiku.
\newblock \emph{Claude-3 Model Card}.

\bibitem[{Bai et~al.(2023{\natexlab{a}})Bai, Bai, Chu, Cui, Dang, Deng, Fan, Ge, Han, Huang et~al.}]{qwen}
Jinze Bai, Shuai Bai, Yunfei Chu, Zeyu Cui, Kai Dang, Xiaodong Deng, Yang Fan, Wenbin Ge, Yu~Han, Fei Huang, et~al. 2023{\natexlab{a}}.
\newblock Qwen technical report.
\newblock \emph{arXiv preprint arXiv:2309.16609}.

\bibitem[{Bai et~al.(2023{\natexlab{b}})Bai, Bai, Yang, Wang, Tan, Wang, Lin, Zhou, and Zhou}]{qwen-vl}
Jinze Bai, Shuai Bai, Shusheng Yang, Shijie Wang, Sinan Tan, Peng Wang, Junyang Lin, Chang Zhou, and Jingren Zhou. 2023{\natexlab{b}}.
\newblock Qwen-vl: A frontier large vision-language model with versatile abilities.
\newblock \emph{arXiv preprint arXiv:2308.12966}.

\bibitem[{Cai et~al.(2024)Cai, Cao, Chen, Chen, Chen, Chen, Chen, Chen, Chen, Chu et~al.}]{internlm2}
Zheng Cai, Maosong Cao, Haojiong Chen, Kai Chen, Keyu Chen, Xin Chen, Xun Chen, Zehui Chen, Zhi Chen, Pei Chu, et~al. 2024.
\newblock Internlm2 technical report.
\newblock \emph{arXiv preprint arXiv:2403.17297}.

\bibitem[{Chen et~al.(2023)Chen, Wu, Wang, Su, Chen, Xing, Muyan, Zhang, Zhu, Lu et~al.}]{chen2023internvl}
Zhe Chen, Jiannan Wu, Wenhai Wang, Weijie Su, Guo Chen, Sen Xing, Zhong Muyan, Qinglong Zhang, Xizhou Zhu, Lewei Lu, et~al. 2023.
\newblock Internvl: Scaling up vision foundation models and aligning for generic visual-linguistic tasks.
\newblock \emph{arXiv preprint arXiv:2312.14238}.

\bibitem[{Huang et~al.(2024)Huang, Wang, Xia, Li, Zou, Xu, Fan, Ye, Chern, Ye, Zhang, Yang, Wu, Wang, Sun, Xiao, Li, Zhou, Chern, Qin, Ma, Su, Liu, Zheng, Zhang, Lin, Qiao, and Liu}]{huang2024olympicarena}
Zhen Huang, Zengzhi Wang, Shijie Xia, Xuefeng Li, Haoyang Zou, Ruijie Xu, Run-Ze Fan, Lyumanshan Ye, Ethan Chern, Yixin Ye, Yikai Zhang, Yuqing Yang, Ting Wu, Binjie Wang, Shichao Sun, Yang Xiao, Yiyuan Li, Fan Zhou, Steffi Chern, Yiwei Qin, Yan Ma, Jiadi Su, Yixiu Liu, Yuxiang Zheng, Shaoting Zhang, Dahua Lin, Yu~Qiao, and Pengfei Liu. 2024.
\newblock \href {http://arxiv.org/abs/2406.12753} {Olympicarena: Benchmarking multi-discipline cognitive reasoning for superintelligent ai}.

\bibitem[{Liu et~al.(2024)Liu, Li, Li, Li, Zhang, Shen, and Lee}]{llava-next}
Haotian Liu, Chunyuan Li, Yuheng Li, Bo~Li, Yuanhan Zhang, Sheng Shen, and Yong~Jae Lee. 2024.
\newblock Llava-next: Improved reasoning, ocr, and world knowledge.

\bibitem[{OpenAI(2024)}]{gpt4o}
OpenAI. 2024.
\newblock \href {https://openai.com/index/hello-gpt-4o/} {Hello gpt-4o}.
\newblock \emph{OpenAI Blog}.

\bibitem[{Reid et~al.(2024)Reid, Savinov, Teplyashin, Lepikhin, Lillicrap, Alayrac, Soricut, Lazaridou, Firat, Schrittwieser et~al.}]{reid2024gemini}
Machel Reid, Nikolay Savinov, Denis Teplyashin, Dmitry Lepikhin, Timothy Lillicrap, Jean-baptiste Alayrac, Radu Soricut, Angeliki Lazaridou, Orhan Firat, Julian Schrittwieser, et~al. 2024.
\newblock Gemini 1.5: Unlocking multimodal understanding across millions of tokens of context.
\newblock \emph{arXiv preprint arXiv:2403.05530}.

\bibitem[{Team et~al.(2023)Team, Anil, Borgeaud, Wu, Alayrac, Yu, Soricut, Schalkwyk, Dai, Hauth et~al.}]{geminiPro}
Gemini Team, Rohan Anil, Sebastian Borgeaud, Yonghui Wu, Jean-Baptiste Alayrac, Jiahui Yu, Radu Soricut, Johan Schalkwyk, Andrew~M Dai, Anja Hauth, et~al. 2023.
\newblock Gemini: a family of highly capable multimodal models.
\newblock \emph{arXiv preprint arXiv:2312.11805}.

\bibitem[{Team et~al.(2024)Team, Georgiev, Lei, Burnell, Bai, Gulati, Tanzer, Vincent, Pan, Wang, Mariooryad, Ding, Geng, Alcober, Frostig, Omernick, Walker, Paduraru, Sorokin, Tacchetti, Gaffney, Daruki, Sercinoglu, Gleicher, Love, Voigtlaender, Jain, Surita, Mohamed, Blevins, Ahn, Zhu, Kawintiranon, Firat, Gu, Zhang, Rahtz, Faruqui, Clay, Gilmer, Co-Reyes, Penchev, Zhu, Morioka, Hui, Haridasan, Campos, Mahdieh, Guo, Hassan, Kilgour, Vezer, Cheng, de~Liedekerke, Goyal, Barham, Strouse, Noury, Adler, Sundararajan, Vikram, Lepikhin, Paganini, Garcia, Yang, Valter, Trebacz, Vodrahalli, Asawaroengchai, Ring, Kalb, Soares, Brahma, Steiner, Yu, Mentzer, He, Gonzalez, Xu, Kaufman, Shafey, Oh, Hennigan, van~den Driessche, Odoom, Lucic, Roelofs, Lall, Marathe, Chan, Ontanon, He, Teplyashin, Lai, Crone, Damoc, Ho, Riedel, Lenc, Yeh, Chowdhery, Xu, Kazemi, Amid, Petrushkina, Swersky, Khodaei, Chen, Larkin, Pinto, Yan, Badia, Patil, Hansen, Orr, Arnold, Grimstad, Dai, Douglas, Sinha, Yadav, Chen, Gribovskaya, Austin,
  Zhao, Patel, Komarek, Austin, Borgeaud, Friso, Goyal, Caine, Cao, Chung, Lamm, Barth-Maron, Kagohara, Olszewska, Chen, Shivakumar, Agarwal, Godhia, Rajwar, Snaider, Dotiwalla, Liu, Barua, Ungureanu, Zhang, Batsaikhan, Wirth, Qin, Danihelka, Doshi, Chadwick, Chen, Jain, Le, Kar, Gurumurthy, Li, Sang, Liu, Lamprou, Munoz, Lintz, Mehta, Howard, Reynolds, Aroyo, Wang, Blanco, Cassirer, Griffith, Das, Lee, Sygnowski, Fisher, Besley, Powell, Ahmed, Paulus, Reitter, Borsos, Joshi, Pope, Hand, Selo, Jain, Sethi, Goel, Makino, May, Yang, Schalkwyk, Butterfield, Hauth, Goldin, Hawkins, Senter, Brin, Woodman, Ritter, Noland, Giang, Bolina, Lee, Blyth, Mackinnon, Reid, Sarvana, Silver, Chen, Wang, Maggiore, Chang, Attaluri, Thornton, Chiu, Bunyan, Levine, Chung, Eltyshev, Si, Lillicrap, Brady, Aggarwal, Wu, Xu, McIlroy, Badola, Sandhu, Moreira, Stokowiec, Hemsley, Li, Tudor, Shyam, Rahimtoroghi, Haykal, Sprechmann, Zhou, Mincu, Li, Addanki, Krishna, Wu, Frechette, Eyal, Dafoe, Lacey, Whang, Avrahami, Zhang, Taropa,
  Lin, Toyama, Rutherford, Sano, Choe, Tomala, Safranek-Shrader, Kassner, Pajarskas, Harvey, Sechrist, Fortunato, Lyu, Elsayed, Kuang, Lottes, Chu, Jia, Chen, Humphreys, Baumli, Tao, Samuel, dos Santos, Andreassen, Rakićević, Grewe, Kumar, Winkler, Caton, Brock, Dalmia, Sheahan, Barr, Miao, Natsev, Devlin, Behbahani, Prost, Sun, Myaskovsky, Pillai, Hurt, Lazaridou, Xiong, Zheng, Pardo, Li, Horgan, Stanton, Ambar, Xia, Lince, Wang, Mustafa, Webson, Lee, Anil, Wicke, Dozat, Sinha, Piqueras, Dabir, Upadhyay, Boral, Hendricks, Fry, Djolonga, Su, Walker, Labanowski, Huang, Misra, Chen, Skerry-Ryan, Singh, Rijhwani, Yu, Castro-Ros, Changpinyo, Datta, Bagri, Hrafnkelsson, Maggioni, Zheng, Sulsky, Hou, Paine, Yang, Riesa, Rogozinska, Marcus, Badawy, Zhang, Wang, Miller, Greer, Sjos, Nova, Zen, Chaabouni, Rosca, Jiang, Chen, Liu, Sainath, Krikun, Polozov, Lespiau, Newlan, Cankara, Kwak, Xu, Chen, Coenen, Meyer, Tsihlas, Ma, Gottweis, Xing, Gu, Miao, Frank, Cankara, Ganapathy, Dasgupta, Hughes-Fitt, Chen, Reid, Rong,
  Fan, van Amersfoort, Zhuang, Cohen, Gu, Mohananey, Ilic, Tobin, Wieting, Bortsova, Thacker, Wang, Caveness, Chiu, Sezener, Kaskasoli, Baker, Millican, Elhawaty, Aisopos, Lebsack, Byrd, Dai, Jia, Wiethoff, Davoodi, Weston, Yagati, Ahuja, Gao, Pundak, Zhang, Azzam, Sim, Caelles, Keeling, Sharma, Swing, Li, Liu, Bostock, Bansal, Nado, Anand, Lipschultz, Karmarkar, Proleev, Ittycheriah, Yeganeh, Polovets, Faust, Sun, Rrustemi, Li, Shivanna, Liu, Welty, Lebron, Baddepudi, Krause, Parisotto, Soricut, Xu, Bloxwich, Johnson, Neyshabur, Mao-Jones, Wang, Ramasesh, Abbas, Guez, Segal, Nguyen, Svensson, Hou, York, Milan, Bridgers, Gworek, Tagliasacchi, Lee-Thorp, Chang, Guseynov, Hartman, Kwong, Zhao, Kashem, Cole, Miech, Tanburn, Phuong, Pavetic, Cevey, Comanescu, Ives, Yang, Du, Li, Zhang, Iinuma, Hu, Roy, Bijwadia, Zhu, Martins, Saputro, Gergely, Zheng, Jia, Antonoglou, Sadovsky, Gu, Bi, Andreev, Samangooei, Khan, Kocisky, Filos, Kumar, Bishop, Yu, Hodkinson, Mittal, Shah, Moufarek, Cheng, Bloniarz, Lee, Pejman,
  Michel, Spencer, Feinberg, Xiong, Savinov, Smith, Shakeri, Tran, Chesus, Bohnet, Tucker, von Glehn, Muir, Mao, Kazawa, Slone, Soparkar, Shrivastava, Cobon-Kerr, Sharman, Pavagadhi, Araya, Misiunas, Ghelani, Laskin, Barker, Li, Briukhov, Houlsby, Glaese, Lakshminarayanan, Schucher, Tang, Collins, Lim, Feng, Recasens, Lai, Magni, Cao, Siddhant, Ashwood, Orbay, Dehghani, Brennan, He, Xu, Gao, Saroufim, Molloy, Wu, Arnold, Chang, Schrittwieser, Buchatskaya, Radpour, Polacek, Giordano, Bapna, Tokumine, Hellendoorn, Sottiaux, Cogan, Severyn, Saleh, Thakoor, Shefey, Qiao, Gaba, yiin Chang, Swanson, Zhang, Lee, Rubenstein, Song, Kwiatkowski, Koop, Kannan, Kao, Schuh, Stjerngren, Ghiasi, Gibson, Vilnis, Yuan, Ferreira, Kamath, Klimenko, Franko, Xiao, Bhattacharya, Patel, Wang, Morris, Strudel, Sharma, Choy, Hashemi, Landon, Finkelstein, Jhakra, Frye, Barnes, Mauger, Daun, Baatarsukh, Tung, Farhan, Michalewski, Viola, de~Chaumont~Quitry, Lan, Hudson, Wang, Fischer, Zheng, White, Dragan, baptiste Alayrac, Ni, Pritzel,
  Iwanicki, Isard, Bulanova, Zilka, Dyer, Sachan, Srinivasan, Muckenhirn, Cai, Mandhane, Tariq, Rae, Wang, Ayoub, FitzGerald, Zhao, Han, Alberti, Garrette, Krishnakumar, Gimenez, Levskaya, Sohn, Matak, Iturrate, Chang, Xiang, Cao, Ranka, Brown, Hutter, Mirrokni, Chen, Yao, Egyed, Galilee, Liechty, Kallakuri, Palmer, Ghemawat, Liu, Tao, Thornton, Green, Jasarevic, Lin, Cotruta, Tan, Fiedel, Yu, Chi, Neitz, Heitkaemper, Sinha, Zhou, Sun, Kaed, Hulse, Mishra, Georgaki, Kudugunta, Farabet, Shafran, Vlasic, Tsitsulin, Ananthanarayanan, Carin, Su, Sun, V, Carvajal, Broder, Comsa, Repina, Wong, Chen, Hawkins, Filonov, Loher, Hirnschall, Wang, Ye, Burns, Cate, Wright, Piccinini, Zhang, Lin, Gog, Kulizhskaya, Sreevatsa, Song, Cobo, Iyer, Tekur, Garrido, Xiao, Kemp, Zheng, Li, Agarwal, Ngani, Goshvadi, Santamaria-Fernandez, Fica, Chen, Gorgolewski, Sun, Garg, Ye, Eslami, Hua, Simon, Joshi, Kim, Tenney, Potluri, Thiet, Yuan, Luisier, Chronopoulou, Scellato, Srinivasan, Chen, Koverkathu, Dalibard, Xu, Saeta, Anderson,
  Sellam, Fernando, Huot, Jung, Varadarajan, Quinn, Raul, Le, Habalov, Clark, Jalan, Bullard, Singhal, Luong, Wang, Rajayogam, Eisenschlos, Jia, Finchelstein, Yakubovich, Balle, Fink, Agarwal, Li, Dvijotham, Pal, Kang, Konzelmann, Beattie, Dousse, Wu, Crocker, Elkind, Jonnalagadda, Lee, Holtmann-Rice, Kallarackal, Liu, Vnukov, Vats, Invernizzi, Jafari, Zhou, Taylor, Prendki, Wu, Eccles, Liu, Kopparapu, Beaufays, Angermueller, Marzoca, Sarcar, Dib, Stanway, Perbet, Trdin, Sterneck, Khorlin, Li, Wu, Goenka, Madras, Goldshtein, Gierke, Zhou, Liu, Liang, White, Li, Singh, Bahargam, Epstein, Basu, Lao, Ozturel, Crous, Zhai, Lu, Tung, Gaur, Walton, Dixon, Zhang, Globerson, Uy, Bolt, Wiles, Nasr, Shumailov, Selvi, Piccinno, Aguilar, McCarthy, Khalman, Shukla, Galic, Carpenter, Villela, Zhang, Richardson, Martens, Bosnjak, Belle, Seibert, Alnahlawi, McWilliams, Singh, Louis, Ding, Popovici, Simicich, Knight, Mehta, Gupta, Shi, Fatehi, Mitrovic, Grills, Pagadora, Petrova, Eisenbud, Zhang, Yates, Mittal, Tripuraneni,
  Assael, Brovelli, Jain, Velimirovic, Akbulut, Mu, Macherey, Kumar, Xu, Qureshi, Comanici, Wiesner, Gong, Ruddock, Bauer, Felt, GP, Arnab, Zelle, Rothfuss, Rosgen, Shenoy, Seybold, Li, Mudigonda, Erdogan, Xia, Simsa, Michi, Yao, Yew, Kan, Caswell, Radebaugh, Elisseeff, Valenzuela, McKinney, Paterson, Cui, Latorre-Chimoto, Kim, Zeng, Durden, Ponnapalli, Sosea, Choquette-Choo, Manyika, Robenek, Vashisht, Pereira, Lam, Velic, Owusu-Afriyie, Lee, Bolukbasi, Parrish, Lu, Park, Venkatraman, Talbert, Rosique, Cheng, Sozanschi, Paszke, Kumar, Austin, Li, Salama, Kim, Dukkipati, Baryshnikov, Kaplanis, Sheng, Chervonyi, Unlu, de~Las~Casas, Askham, Tunyasuvunakool, Gimeno, Poder, Kwak, Miecnikowski, Mirrokni, Dimitriev, Parisi, Liu, Tsai, Shevlane, Kouridi, Garmon, Goedeckemeyer, Brown, Vijayakumar, Elqursh, Jazayeri, Huang, Carthy, Hoover, Kim, Kumar, Chen, Biles, Bingham, Rosen, Wang, Tan, Engel, Pongetti, de~Cesare, Hwang, Yu, Pullman, Narayanan, Levin, Gopal, Li, Aharoni, Trinh, Lo, Casagrande, Vij, Matthey,
  Ramadhana, Matthews, Carey, Johnson, Goranova, Shah, Ashraf, Dasgupta, Larsen, Wang, Vuyyuru, Jiang, Ijazi, Osawa, Smith, Boppana, Bilal, Koizumi, Xu, Altun, Shabat, Bariach, Korchemniy, Choo, Ronneberger, Iwuanyanwu, Zhao, Soergel, Hsieh, Cai, Iqbal, Sundermeyer, Chen, Bursztein, Malaviya, Biadsy, Shroff, Dhillon, Latkar, Dyer, Forbes, Nicosia, Nikolaev, Greene, Georgiev, Wang, Martin, Sedghi, Zhang, Banzal, Fritz, Rao, Wang, Zhang, Patraucean, Du, Mordatch, Jurin, Liu, Dubey, Mohan, Nowakowski, Ion, Wei, Tojo, Raad, Hudson, Keshava, Agrawal, Ramirez, Wu, Nguyen, Liu, Sewak, Petrini, Choi, Philips, Wang, Bica, Garg, Wilkiewicz, Agrawal, Li, Guo, Xue, Shaik, Leach, Khan, Wiesinger, Jerome, Chakladar, Wang, Ornduff, Abu, Ghaffarkhah, Wainwright, Cortes, Liu, Maynez, Petrov, Wu, Hassabis, Kavukcuoglu, Dean, and Vinyals}]{geminiteam2024gemini}
Gemini Team, Petko Georgiev, Ving~Ian Lei, Ryan Burnell, Libin Bai, Anmol Gulati, Garrett Tanzer, Damien Vincent, Zhufeng Pan, Shibo Wang, Soroosh Mariooryad, Yifan Ding, Xinyang Geng, Fred Alcober, Roy Frostig, Mark Omernick, Lexi Walker, Cosmin Paduraru, Christina Sorokin, Andrea Tacchetti, Colin Gaffney, Samira Daruki, Olcan Sercinoglu, Zach Gleicher, Juliette Love, Paul Voigtlaender, Rohan Jain, Gabriela Surita, Kareem Mohamed, Rory Blevins, Junwhan Ahn, Tao Zhu, Kornraphop Kawintiranon, Orhan Firat, Yiming Gu, Yujing Zhang, Matthew Rahtz, Manaal Faruqui, Natalie Clay, Justin Gilmer, JD~Co-Reyes, Ivo Penchev, Rui Zhu, Nobuyuki Morioka, Kevin Hui, Krishna Haridasan, Victor Campos, Mahdis Mahdieh, Mandy Guo, Samer Hassan, Kevin Kilgour, Arpi Vezer, Heng-Tze Cheng, Raoul de~Liedekerke, Siddharth Goyal, Paul Barham, DJ~Strouse, Seb Noury, Jonas Adler, Mukund Sundararajan, Sharad Vikram, Dmitry Lepikhin, Michela Paganini, Xavier Garcia, Fan Yang, Dasha Valter, Maja Trebacz, Kiran Vodrahalli, Chulayuth
  Asawaroengchai, Roman Ring, Norbert Kalb, Livio~Baldini Soares, Siddhartha Brahma, David Steiner, Tianhe Yu, Fabian Mentzer, Antoine He, Lucas Gonzalez, Bibo Xu, Raphael~Lopez Kaufman, Laurent~El Shafey, Junhyuk Oh, Tom Hennigan, George van~den Driessche, Seth Odoom, Mario Lucic, Becca Roelofs, Sid Lall, Amit Marathe, Betty Chan, Santiago Ontanon, Luheng He, Denis Teplyashin, Jonathan Lai, Phil Crone, Bogdan Damoc, Lewis Ho, Sebastian Riedel, Karel Lenc, Chih-Kuan Yeh, Aakanksha Chowdhery, Yang Xu, Mehran Kazemi, Ehsan Amid, Anastasia Petrushkina, Kevin Swersky, Ali Khodaei, Gowoon Chen, Chris Larkin, Mario Pinto, Geng Yan, Adria~Puigdomenech Badia, Piyush Patil, Steven Hansen, Dave Orr, Sebastien M.~R. Arnold, Jordan Grimstad, Andrew Dai, Sholto Douglas, Rishika Sinha, Vikas Yadav, Xi~Chen, Elena Gribovskaya, Jacob Austin, Jeffrey Zhao, Kaushal Patel, Paul Komarek, Sophia Austin, Sebastian Borgeaud, Linda Friso, Abhimanyu Goyal, Ben Caine, Kris Cao, Da-Woon Chung, Matthew Lamm, Gabe Barth-Maron, Thais
  Kagohara, Kate Olszewska, Mia Chen, Kaushik Shivakumar, Rishabh Agarwal, Harshal Godhia, Ravi Rajwar, Javier Snaider, Xerxes Dotiwalla, Yuan Liu, Aditya Barua, Victor Ungureanu, Yuan Zhang, Bat-Orgil Batsaikhan, Mateo Wirth, James Qin, Ivo Danihelka, Tulsee Doshi, Martin Chadwick, Jilin Chen, Sanil Jain, Quoc Le, Arjun Kar, Madhu Gurumurthy, Cheng Li, Ruoxin Sang, Fangyu Liu, Lampros Lamprou, Rich Munoz, Nathan Lintz, Harsh Mehta, Heidi Howard, Malcolm Reynolds, Lora Aroyo, Quan Wang, Lorenzo Blanco, Albin Cassirer, Jordan Griffith, Dipanjan Das, Stephan Lee, Jakub Sygnowski, Zach Fisher, James Besley, Richard Powell, Zafarali Ahmed, Dominik Paulus, David Reitter, Zalan Borsos, Rishabh Joshi, Aedan Pope, Steven Hand, Vittorio Selo, Vihan Jain, Nikhil Sethi, Megha Goel, Takaki Makino, Rhys May, Zhen Yang, Johan Schalkwyk, Christina Butterfield, Anja Hauth, Alex Goldin, Will Hawkins, Evan Senter, Sergey Brin, Oliver Woodman, Marvin Ritter, Eric Noland, Minh Giang, Vijay Bolina, Lisa Lee, Tim Blyth, Ian
  Mackinnon, Machel Reid, Obaid Sarvana, David Silver, Alexander Chen, Lily Wang, Loren Maggiore, Oscar Chang, Nithya Attaluri, Gregory Thornton, Chung-Cheng Chiu, Oskar Bunyan, Nir Levine, Timothy Chung, Evgenii Eltyshev, Xiance Si, Timothy Lillicrap, Demetra Brady, Vaibhav Aggarwal, Boxi Wu, Yuanzhong Xu, Ross McIlroy, Kartikeya Badola, Paramjit Sandhu, Erica Moreira, Wojciech Stokowiec, Ross Hemsley, Dong Li, Alex Tudor, Pranav Shyam, Elahe Rahimtoroghi, Salem Haykal, Pablo Sprechmann, Xiang Zhou, Diana Mincu, Yujia Li, Ravi Addanki, Kalpesh Krishna, Xiao Wu, Alexandre Frechette, Matan Eyal, Allan Dafoe, Dave Lacey, Jay Whang, Thi Avrahami, Ye~Zhang, Emanuel Taropa, Hanzhao Lin, Daniel Toyama, Eliza Rutherford, Motoki Sano, HyunJeong Choe, Alex Tomala, Chalence Safranek-Shrader, Nora Kassner, Mantas Pajarskas, Matt Harvey, Sean Sechrist, Meire Fortunato, Christina Lyu, Gamaleldin Elsayed, Chenkai Kuang, James Lottes, Eric Chu, Chao Jia, Chih-Wei Chen, Peter Humphreys, Kate Baumli, Connie Tao, Rajkumar
  Samuel, Cicero~Nogueira dos Santos, Anders Andreassen, Nemanja Rakićević, Dominik Grewe, Aviral Kumar, Stephanie Winkler, Jonathan Caton, Andrew Brock, Sid Dalmia, Hannah Sheahan, Iain Barr, Yingjie Miao, Paul Natsev, Jacob Devlin, Feryal Behbahani, Flavien Prost, Yanhua Sun, Artiom Myaskovsky, Thanumalayan~Sankaranarayana Pillai, Dan Hurt, Angeliki Lazaridou, Xi~Xiong, Ce~Zheng, Fabio Pardo, Xiaowei Li, Dan Horgan, Joe Stanton, Moran Ambar, Fei Xia, Alejandro Lince, Mingqiu Wang, Basil Mustafa, Albert Webson, Hyo Lee, Rohan Anil, Martin Wicke, Timothy Dozat, Abhishek Sinha, Enrique Piqueras, Elahe Dabir, Shyam Upadhyay, Anudhyan Boral, Lisa~Anne Hendricks, Corey Fry, Josip Djolonga, Yi~Su, Jake Walker, Jane Labanowski, Ronny Huang, Vedant Misra, Jeremy Chen, RJ~Skerry-Ryan, Avi Singh, Shruti Rijhwani, Dian Yu, Alex Castro-Ros, Beer Changpinyo, Romina Datta, Sumit Bagri, Arnar~Mar Hrafnkelsson, Marcello Maggioni, Daniel Zheng, Yury Sulsky, Shaobo Hou, Tom~Le Paine, Antoine Yang, Jason Riesa, Dominika
  Rogozinska, Dror Marcus, Dalia~El Badawy, Qiao Zhang, Luyu Wang, Helen Miller, Jeremy Greer, Lars~Lowe Sjos, Azade Nova, Heiga Zen, Rahma Chaabouni, Mihaela Rosca, Jiepu Jiang, Charlie Chen, Ruibo Liu, Tara Sainath, Maxim Krikun, Alex Polozov, Jean-Baptiste Lespiau, Josh Newlan, Zeyncep Cankara, Soo Kwak, Yunhan Xu, Phil Chen, Andy Coenen, Clemens Meyer, Katerina Tsihlas, Ada Ma, Juraj Gottweis, Jinwei Xing, Chenjie Gu, Jin Miao, Christian Frank, Zeynep Cankara, Sanjay Ganapathy, Ishita Dasgupta, Steph Hughes-Fitt, Heng Chen, David Reid, Keran Rong, Hongmin Fan, Joost van Amersfoort, Vincent Zhuang, Aaron Cohen, Shixiang~Shane Gu, Anhad Mohananey, Anastasija Ilic, Taylor Tobin, John Wieting, Anna Bortsova, Phoebe Thacker, Emma Wang, Emily Caveness, Justin Chiu, Eren Sezener, Alex Kaskasoli, Steven Baker, Katie Millican, Mohamed Elhawaty, Kostas Aisopos, Carl Lebsack, Nathan Byrd, Hanjun Dai, Wenhao Jia, Matthew Wiethoff, Elnaz Davoodi, Albert Weston, Lakshman Yagati, Arun Ahuja, Isabel Gao, Golan Pundak,
  Susan Zhang, Michael Azzam, Khe~Chai Sim, Sergi Caelles, James Keeling, Abhanshu Sharma, Andy Swing, YaGuang Li, Chenxi Liu, Carrie~Grimes Bostock, Yamini Bansal, Zachary Nado, Ankesh Anand, Josh Lipschultz, Abhijit Karmarkar, Lev Proleev, Abe Ittycheriah, Soheil~Hassas Yeganeh, George Polovets, Aleksandra Faust, Jiao Sun, Alban Rrustemi, Pen Li, Rakesh Shivanna, Jeremiah Liu, Chris Welty, Federico Lebron, Anirudh Baddepudi, Sebastian Krause, Emilio Parisotto, Radu Soricut, Zheng Xu, Dawn Bloxwich, Melvin Johnson, Behnam Neyshabur, Justin Mao-Jones, Renshen Wang, Vinay Ramasesh, Zaheer Abbas, Arthur Guez, Constant Segal, Duc~Dung Nguyen, James Svensson, Le~Hou, Sarah York, Kieran Milan, Sophie Bridgers, Wiktor Gworek, Marco Tagliasacchi, James Lee-Thorp, Michael Chang, Alexey Guseynov, Ale~Jakse Hartman, Michael Kwong, Ruizhe Zhao, Sheleem Kashem, Elizabeth Cole, Antoine Miech, Richard Tanburn, Mary Phuong, Filip Pavetic, Sebastien Cevey, Ramona Comanescu, Richard Ives, Sherry Yang, Cosmo Du, Bo~Li, Zizhao
  Zhang, Mariko Iinuma, Clara~Huiyi Hu, Aurko Roy, Shaan Bijwadia, Zhenkai Zhu, Danilo Martins, Rachel Saputro, Anita Gergely, Steven Zheng, Dawei Jia, Ioannis Antonoglou, Adam Sadovsky, Shane Gu, Yingying Bi, Alek Andreev, Sina Samangooei, Mina Khan, Tomas Kocisky, Angelos Filos, Chintu Kumar, Colton Bishop, Adams Yu, Sarah Hodkinson, Sid Mittal, Premal Shah, Alexandre Moufarek, Yong Cheng, Adam Bloniarz, Jaehoon Lee, Pedram Pejman, Paul Michel, Stephen Spencer, Vladimir Feinberg, Xuehan Xiong, Nikolay Savinov, Charlotte Smith, Siamak Shakeri, Dustin Tran, Mary Chesus, Bernd Bohnet, George Tucker, Tamara von Glehn, Carrie Muir, Yiran Mao, Hideto Kazawa, Ambrose Slone, Kedar Soparkar, Disha Shrivastava, James Cobon-Kerr, Michael Sharman, Jay Pavagadhi, Carlos Araya, Karolis Misiunas, Nimesh Ghelani, Michael Laskin, David Barker, Qiujia Li, Anton Briukhov, Neil Houlsby, Mia Glaese, Balaji Lakshminarayanan, Nathan Schucher, Yunhao Tang, Eli Collins, Hyeontaek Lim, Fangxiaoyu Feng, Adria Recasens, Guangda Lai,
  Alberto Magni, Nicola~De Cao, Aditya Siddhant, Zoe Ashwood, Jordi Orbay, Mostafa Dehghani, Jenny Brennan, Yifan He, Kelvin Xu, Yang Gao, Carl Saroufim, James Molloy, Xinyi Wu, Seb Arnold, Solomon Chang, Julian Schrittwieser, Elena Buchatskaya, Soroush Radpour, Martin Polacek, Skye Giordano, Ankur Bapna, Simon Tokumine, Vincent Hellendoorn, Thibault Sottiaux, Sarah Cogan, Aliaksei Severyn, Mohammad Saleh, Shantanu Thakoor, Laurent Shefey, Siyuan Qiao, Meenu Gaba, Shuo yiin Chang, Craig Swanson, Biao Zhang, Benjamin Lee, Paul~Kishan Rubenstein, Gan Song, Tom Kwiatkowski, Anna Koop, Ajay Kannan, David Kao, Parker Schuh, Axel Stjerngren, Golnaz Ghiasi, Gena Gibson, Luke Vilnis, Ye~Yuan, Felipe~Tiengo Ferreira, Aishwarya Kamath, Ted Klimenko, Ken Franko, Kefan Xiao, Indro Bhattacharya, Miteyan Patel, Rui Wang, Alex Morris, Robin Strudel, Vivek Sharma, Peter Choy, Sayed~Hadi Hashemi, Jessica Landon, Mara Finkelstein, Priya Jhakra, Justin Frye, Megan Barnes, Matthew Mauger, Dennis Daun, Khuslen Baatarsukh, Matthew
  Tung, Wael Farhan, Henryk Michalewski, Fabio Viola, Felix de~Chaumont~Quitry, Charline~Le Lan, Tom Hudson, Qingze Wang, Felix Fischer, Ivy Zheng, Elspeth White, Anca Dragan, Jean baptiste Alayrac, Eric Ni, Alexander Pritzel, Adam Iwanicki, Michael Isard, Anna Bulanova, Lukas Zilka, Ethan Dyer, Devendra Sachan, Srivatsan Srinivasan, Hannah Muckenhirn, Honglong Cai, Amol Mandhane, Mukarram Tariq, Jack~W. Rae, Gary Wang, Kareem Ayoub, Nicholas FitzGerald, Yao Zhao, Woohyun Han, Chris Alberti, Dan Garrette, Kashyap Krishnakumar, Mai Gimenez, Anselm Levskaya, Daniel Sohn, Josip Matak, Inaki Iturrate, Michael~B. Chang, Jackie Xiang, Yuan Cao, Nishant Ranka, Geoff Brown, Adrian Hutter, Vahab Mirrokni, Nanxin Chen, Kaisheng Yao, Zoltan Egyed, Francois Galilee, Tyler Liechty, Praveen Kallakuri, Evan Palmer, Sanjay Ghemawat, Jasmine Liu, David Tao, Chloe Thornton, Tim Green, Mimi Jasarevic, Sharon Lin, Victor Cotruta, Yi-Xuan Tan, Noah Fiedel, Hongkun Yu, Ed~Chi, Alexander Neitz, Jens Heitkaemper, Anu Sinha, Denny
  Zhou, Yi~Sun, Charbel Kaed, Brice Hulse, Swaroop Mishra, Maria Georgaki, Sneha Kudugunta, Clement Farabet, Izhak Shafran, Daniel Vlasic, Anton Tsitsulin, Rajagopal Ananthanarayanan, Alen Carin, Guolong Su, Pei Sun, Shashank V, Gabriel Carvajal, Josef Broder, Iulia Comsa, Alena Repina, William Wong, Warren~Weilun Chen, Peter Hawkins, Egor Filonov, Lucia Loher, Christoph Hirnschall, Weiyi Wang, Jingchen Ye, Andrea Burns, Hardie Cate, Diana~Gage Wright, Federico Piccinini, Lei Zhang, Chu-Cheng Lin, Ionel Gog, Yana Kulizhskaya, Ashwin Sreevatsa, Shuang Song, Luis~C. Cobo, Anand Iyer, Chetan Tekur, Guillermo Garrido, Zhuyun Xiao, Rupert Kemp, Huaixiu~Steven Zheng, Hui Li, Ananth Agarwal, Christel Ngani, Kati Goshvadi, Rebeca Santamaria-Fernandez, Wojciech Fica, Xinyun Chen, Chris Gorgolewski, Sean Sun, Roopal Garg, Xinyu Ye, S.~M.~Ali Eslami, Nan Hua, Jon Simon, Pratik Joshi, Yelin Kim, Ian Tenney, Sahitya Potluri, Lam~Nguyen Thiet, Quan Yuan, Florian Luisier, Alexandra Chronopoulou, Salvatore Scellato, Praveen
  Srinivasan, Minmin Chen, Vinod Koverkathu, Valentin Dalibard, Yaming Xu, Brennan Saeta, Keith Anderson, Thibault Sellam, Nick Fernando, Fantine Huot, Junehyuk Jung, Mani Varadarajan, Michael Quinn, Amit Raul, Maigo Le, Ruslan Habalov, Jon Clark, Komal Jalan, Kalesha Bullard, Achintya Singhal, Thang Luong, Boyu Wang, Sujeevan Rajayogam, Julian Eisenschlos, Johnson Jia, Daniel Finchelstein, Alex Yakubovich, Daniel Balle, Michael Fink, Sameer Agarwal, Jing Li, Dj~Dvijotham, Shalini Pal, Kai Kang, Jaclyn Konzelmann, Jennifer Beattie, Olivier Dousse, Diane Wu, Remi Crocker, Chen Elkind, Siddhartha~Reddy Jonnalagadda, Jong Lee, Dan Holtmann-Rice, Krystal Kallarackal, Rosanne Liu, Denis Vnukov, Neera Vats, Luca Invernizzi, Mohsen Jafari, Huanjie Zhou, Lilly Taylor, Jennifer Prendki, Marcus Wu, Tom Eccles, Tianqi Liu, Kavya Kopparapu, Francoise Beaufays, Christof Angermueller, Andreea Marzoca, Shourya Sarcar, Hilal Dib, Jeff Stanway, Frank Perbet, Nejc Trdin, Rachel Sterneck, Andrey Khorlin, Dinghua Li, Xihui Wu,
  Sonam Goenka, David Madras, Sasha Goldshtein, Willi Gierke, Tong Zhou, Yaxin Liu, Yannie Liang, Anais White, Yunjie Li, Shreya Singh, Sanaz Bahargam, Mark Epstein, Sujoy Basu, Li~Lao, Adnan Ozturel, Carl Crous, Alex Zhai, Han Lu, Zora Tung, Neeraj Gaur, Alanna Walton, Lucas Dixon, Ming Zhang, Amir Globerson, Grant Uy, Andrew Bolt, Olivia Wiles, Milad Nasr, Ilia Shumailov, Marco Selvi, Francesco Piccinno, Ricardo Aguilar, Sara McCarthy, Misha Khalman, Mrinal Shukla, Vlado Galic, John Carpenter, Kevin Villela, Haibin Zhang, Harry Richardson, James Martens, Matko Bosnjak, Shreyas~Rammohan Belle, Jeff Seibert, Mahmoud Alnahlawi, Brian McWilliams, Sankalp Singh, Annie Louis, Wen Ding, Dan Popovici, Lenin Simicich, Laura Knight, Pulkit Mehta, Nishesh Gupta, Chongyang Shi, Saaber Fatehi, Jovana Mitrovic, Alex Grills, Joseph Pagadora, Dessie Petrova, Danielle Eisenbud, Zhishuai Zhang, Damion Yates, Bhavishya Mittal, Nilesh Tripuraneni, Yannis Assael, Thomas Brovelli, Prateek Jain, Mihajlo Velimirovic, Canfer
  Akbulut, Jiaqi Mu, Wolfgang Macherey, Ravin Kumar, Jun Xu, Haroon Qureshi, Gheorghe Comanici, Jeremy Wiesner, Zhitao Gong, Anton Ruddock, Matthias Bauer, Nick Felt, Anirudh GP, Anurag Arnab, Dustin Zelle, Jonas Rothfuss, Bill Rosgen, Ashish Shenoy, Bryan Seybold, Xinjian Li, Jayaram Mudigonda, Goker Erdogan, Jiawei Xia, Jiri Simsa, Andrea Michi, Yi~Yao, Christopher Yew, Steven Kan, Isaac Caswell, Carey Radebaugh, Andre Elisseeff, Pedro Valenzuela, Kay McKinney, Kim Paterson, Albert Cui, Eri Latorre-Chimoto, Solomon Kim, William Zeng, Ken Durden, Priya Ponnapalli, Tiberiu Sosea, Christopher~A. Choquette-Choo, James Manyika, Brona Robenek, Harsha Vashisht, Sebastien Pereira, Hoi Lam, Marko Velic, Denese Owusu-Afriyie, Katherine Lee, Tolga Bolukbasi, Alicia Parrish, Shawn Lu, Jane Park, Balaji Venkatraman, Alice Talbert, Lambert Rosique, Yuchung Cheng, Andrei Sozanschi, Adam Paszke, Praveen Kumar, Jessica Austin, Lu~Li, Khalid Salama, Wooyeol Kim, Nandita Dukkipati, Anthony Baryshnikov, Christos Kaplanis,
  XiangHai Sheng, Yuri Chervonyi, Caglar Unlu, Diego de~Las~Casas, Harry Askham, Kathryn Tunyasuvunakool, Felix Gimeno, Siim Poder, Chester Kwak, Matt Miecnikowski, Vahab Mirrokni, Alek Dimitriev, Aaron Parisi, Dangyi Liu, Tomy Tsai, Toby Shevlane, Christina Kouridi, Drew Garmon, Adrian Goedeckemeyer, Adam~R. Brown, Anitha Vijayakumar, Ali Elqursh, Sadegh Jazayeri, Jin Huang, Sara~Mc Carthy, Jay Hoover, Lucy Kim, Sandeep Kumar, Wei Chen, Courtney Biles, Garrett Bingham, Evan Rosen, Lisa Wang, Qijun Tan, David Engel, Francesco Pongetti, Dario de~Cesare, Dongseong Hwang, Lily Yu, Jennifer Pullman, Srini Narayanan, Kyle Levin, Siddharth Gopal, Megan Li, Asaf Aharoni, Trieu Trinh, Jessica Lo, Norman Casagrande, Roopali Vij, Loic Matthey, Bramandia Ramadhana, Austin Matthews, CJ~Carey, Matthew Johnson, Kremena Goranova, Rohin Shah, Shereen Ashraf, Kingshuk Dasgupta, Rasmus Larsen, Yicheng Wang, Manish~Reddy Vuyyuru, Chong Jiang, Joana Ijazi, Kazuki Osawa, Celine Smith, Ramya~Sree Boppana, Taylan Bilal, Yuma
  Koizumi, Ying Xu, Yasemin Altun, Nir Shabat, Ben Bariach, Alex Korchemniy, Kiam Choo, Olaf Ronneberger, Chimezie Iwuanyanwu, Shubin Zhao, David Soergel, Cho-Jui Hsieh, Irene Cai, Shariq Iqbal, Martin Sundermeyer, Zhe Chen, Elie Bursztein, Chaitanya Malaviya, Fadi Biadsy, Prakash Shroff, Inderjit Dhillon, Tejasi Latkar, Chris Dyer, Hannah Forbes, Massimo Nicosia, Vitaly Nikolaev, Somer Greene, Marin Georgiev, Pidong Wang, Nina Martin, Hanie Sedghi, John Zhang, Praseem Banzal, Doug Fritz, Vikram Rao, Xuezhi Wang, Jiageng Zhang, Viorica Patraucean, Dayou Du, Igor Mordatch, Ivan Jurin, Lewis Liu, Ayush Dubey, Abhi Mohan, Janek Nowakowski, Vlad-Doru Ion, Nan Wei, Reiko Tojo, Maria~Abi Raad, Drew~A. Hudson, Vaishakh Keshava, Shubham Agrawal, Kevin Ramirez, Zhichun Wu, Hoang Nguyen, Ji~Liu, Madhavi Sewak, Bryce Petrini, DongHyun Choi, Ivan Philips, Ziyue Wang, Ioana Bica, Ankush Garg, Jarek Wilkiewicz, Priyanka Agrawal, Xiaowei Li, Danhao Guo, Emily Xue, Naseer Shaik, Andrew Leach, Sadh~MNM Khan, Julia Wiesinger,
  Sammy Jerome, Abhishek Chakladar, Alek~Wenjiao Wang, Tina Ornduff, Folake Abu, Alireza Ghaffarkhah, Marcus Wainwright, Mario Cortes, Frederick Liu, Joshua Maynez, Slav Petrov, Yonghui Wu, Demis Hassabis, Koray Kavukcuoglu, Jeffrey Dean, and Oriol Vinyals. 2024.
\newblock \href {http://arxiv.org/abs/2403.05530} {Gemini 1.5: Unlocking multimodal understanding across millions of tokens of context}.

\bibitem[{Young et~al.(2024)Young, Chen, Li, Huang, Zhang, Zhang, Li, Zhu, Chen, Chang et~al.}]{yi}
Alex Young, Bei Chen, Chao Li, Chengen Huang, Ge~Zhang, Guanwei Zhang, Heng Li, Jiangcheng Zhu, Jianqun Chen, Jing Chang, et~al. 2024.
\newblock Yi: Open foundation models by 01. ai.
\newblock \emph{arXiv preprint arXiv:2403.04652}.

\end{thebibliography}
\bibliographystyle{acl_natbib}

\end{document}